\documentclass[runningheads]{llncs}
\usepackage{graphicx}
\usepackage{xcolor}
\usepackage{csquotes}
\usepackage{mdframed}
\newcommand{\reasonx}[1]{{\scshape reasonx}}

\usepackage{hyperref}

\begin{document}
\title{Reason to explain: Interactive contrastive explanations ({\scshape REASONX})
}

\titlerunning{Reason to explain: \reasonx{}}
%
\author{Laura State\inst{1,2}\orcidID{0000-0001-8084-5297} \and
Salvatore Ruggieri\inst{1}\orcidID{0000-0002-1917-6087} \and
Franco Turini\inst{1}\orcidID{0000-0001-6789-5476}}
\authorrunning{L. State et al.}

%
\institute{University of Pisa, Pisa, Italy \and
Scuola Normale Superiore, Pisa, Italy
}

\maketitle              

\begin{abstract}

Many high-performing machine learning models are not interpretable. As they are increasingly used in decision scenarios that can critically affect individuals, it is necessary to develop tools to better understand their outputs.
Popular explanation methods include contrastive explanations. However, they suffer several shortcomings, among others an insufficient incorporation of background knowledge, and a lack of interactivity.
While (dialogue-like) interactivity is important to better communicate an explanation, background knowledge has the potential to significantly improve their quality, e.g., by adapting the explanation to the needs of the end-user. 

To close this gap, we present \reasonx{}, an explanation tool based on Constraint Logic Programming (CLP).
\reasonx{} provides interactive contrastive explanations that can be augmented by background knowledge, 
and allows to operate under a setting of under-specified information, leading to increased flexibility in the provided explanations.
\reasonx{} computes factual and constrative decision rules, as well as closest constrative examples.
It provides explanations for decision trees, which can be the ML models under analysis, or global/local surrogate models of any ML model.

While the core part of \reasonx{} is built on CLP, we also provide a program layer that allows to compute the explanations via Python, making the tool accessible to a wider audience.
We illustrate the capability of \reasonx{} on a synthetic data set, and on a a well-developed example in the credit domain.
In both cases, we can show how \reasonx{} can be flexibly used and tailored to the needs of the user.

\end{abstract}
\section{Introduction}

Contrastive (counterfactual) explanations (CEs)
\footnote{In this paper, we refer to \textit{contrastive} over counterfactual explanations, in order to avoid confusion with the concept of counterfactuals as understood in causality, and following~\cite{DBLP:journals/ai/Miller19}. 
}
are a popular method to provide insights into not interpretable machine learning (ML) models.
Significant efforts are made to provide CEs for different data types such as tabular data~\cite{DBLP:journals/corr/abs-1711-00399}, images~\cite{DBLP:conf/iclr/ChangCGD19} or text~\cite{DBLP:conf/acl/WuRHW20}, 
based on numeric~\cite{DBLP:conf/fat/MothilalST20,DBLP:journals/corr/abs-1711-00399}, causal~\cite{DBLP:conf/fat/KarimiSV21} or other approaches~\cite{DBLP:conf/fat/Russell19,DBLP:conf/fat/UstunSL19}. 
However, some open challenges remain, such as the integration of background knowledge~\cite{DBLP:journals/corr/abs-2105-10172,DBLP:conf/iclp/State21}, and interactivity~\cite{DBLP:journals/ai/Miller19,DBLP:journals/corr/abs-1712-00547,DBLP:journals/cacm/WeldB19}.

To tackle these shortcomings, we present \reasonx{} (REASON to eXplain).
We display an exemplary dialogue between a fictional end-user and \reasonx{} below. 
The dialogue is situated in the context of a credit application example, therefore the user of the tool is a natural person whose credit application has been rejected by an Automated Decision-Making (ADM) system.
Please note that while the information content of the displayed dialogue is exactly what \reasonx{} can provide, we enhanced the dialogue to mimic better a realistic interaction by reporting questions and answers in natural language.
\footnote{Adding a layer that implements a natural language communication between the user and \reasonx{} is left for future work.}

\begin{small}
\begin{mdframed}

\texttt{USER}: I want to understand the decision better. Can I see the rule that led to the denial of my credit application? \\
\texttt{REASONX}: Your credit application was rejected, because your income is lower than 60,000 EUR/year, and you still have to pay back the lease of your car. \\
\texttt{USER}: Ok. Can you present me two different options that will lead to a change of the decision outcome? Please take into consideration that I need a credit of at least 10,000 EUR. 
{I would like to see options that require as little change as necessary.} \\
\texttt{REASONX}: You have the following two options: 
you pay back the lease on the car, or you increase your age by 10 years (from 35 to 45 years). \\
\texttt{USER}: The second option presented is a bit strange. I am wondering whether this is something salient in the model. 
Can I please see the options to obtain credit for an individual with the same properties as me, for a credit of at least 10,000 EUR, but with the feature age at 35 years or less (i.e., young applicant), instead of fixed? \\
\texttt{REASONX}: For the given profile, the credit is always rejected. \\
\texttt{USER}: Can you please show how to reverse the decision, under as few changes as possible, for the specified profile? \\
\texttt{REASONX}: Credit can be obtained, if the feature age is set to higher than 35 years. \\
\texttt{USER}: This is interesting and worth investigating further. There could be bias w.r.t. the age of the person that applies for credit.

\end{mdframed}
\end{small}

This dialogue illustrates the capabilities of \reasonx{}. Our tool addresses the following points:

\paragraph{Background knowledge.}
Adding background knowledge to an explanation has the potential to significantly improve its quality, but it is seldom offered~\cite{DBLP:journals/corr/abs-2105-10172,DBLP:conf/iclp/State21}. 
Ignoring this knowledge is not necessarily wrong, but can lead to contrastive explanations that ignore the needs of the data subject under decision, or CEs that do not fit the reality of our world - this depends closely on the purpose of the explanation.
An example of such knowledge in the credit application example is the minimum credit amount (\enquote{a credit of at least 10,000 EUR}, see dialogue above).
For \reasonx{}, we rely on background knowledge in the form of linear constraints over the features. 

\paragraph{Interactivity.}
Interactivity is a property of an explanation that is important, if the explanations should be communicated successfully to the end-user~\cite{DBLP:journals/ai/Miller19,DBLP:journals/corr/abs-1712-00547,DBLP:journals/cacm/WeldB19}. 
However, most eXplainabe AI (XAI) tools do not account for this.
Interactivity arises naturally in \reasonx{}: the end-user can flexibly query the explanation tool, choosing answers that best fit her questions, adding and removing constraints, and thereby building her own, personalized explanation.

\paragraph{Under-specified information.}
\reasonx{} allows computing CEs under under-speci\-fied information. 
For example, it is not necessary to fix all features of the data instance of interest to compute a matching CE but provide only bounds (\enquote{feature age at 35 years or less}, see dialogue above).
This property is implicitly in the use of CLP, and leads to a wider flexibility in provided explanations.
Further, there is a {loose} connection to the notion of \textit{group contrastive explanations}, which have been shown to be beneficial to the end-user via user studies~\cite{DBLP:journals/corr/abs-2303-09297}.

\subsection{Contributions and Structure of the Paper}

With this paper, we make the following contributions

\begin{itemize}
    \item \textit{Explanation tool.} We propose \reasonx{}, a novel explanation tool that provides contrastive interactive explanations, that can incorporate background knowledge, and that works under under-specified information.
    To the best of our knowledge, this is one of the first tools using constraint logic programming to generate contrastive explanations.
    \item \textit{Synthetic data set illustration.} 
    A first illustration of \reasonx{} is based on a well-defined synthetic data set. We demonstrate step-by-step how \reasonx{} operates under different (constraint) settings, including some graphical representations.
    \item \textit{Credit application example.} We demonstrate the capabilities of \reasonx{} on the Adult Income data set.
    This data set can be used as an approximation to determine the income of a natural person, we assume here that this approximation relates directly to a decision about credit allocation.
    We provide this demonstration together with an in-depth discussion of the relevant context (credit domain).
    We chose this context, as ADM systems for credit applications are an important and recurring topic. 
    Further, as they are classified as \enquote{high-risk} according to the proposed AI Act of the European Union~\cite{Commission2021}, they have to be considered particularly carefully. 
\end{itemize}

This paper is structured as follows. In Sec. \ref{sec:related_work}, we discuss background and related work. In Sec. \ref{sec:reasonx}, we introduce \reasonx{}, followed by the illustration on synthetic data in Sec. \ref{sec:synthetic_dataset} and the credit application example in Sec. \ref{sec:credit_use_case}.
We close our paper by discussing limitations of this work in Sec. \ref{sec:limitations}, and a conclusion in Sec. \ref{sec:conclusion}.

We provide our code (\reasonx{} and experiments) via a public repository. \footnote{\href{https://github.com/lstate/REASONX}{https://github.com/lstate/REASONX}}
In-depth theory and implementation details of \reasonx{} will be provided in a companion paper.

\section{Background and Related Work}
\label{sec:related_work}

\subsection{Explanations}

Explanations for opaque ML models (``black box models'') can be divided into two categories \cite{DBLP:journals/csur/GuidottiMRTGP19}: \textit{global} methods, explaining the full model at once, or \textit{local} methods, producing explanations that are only valid for a single data instance (data subject). Further, we can distinguish between \textit{model-agnostic} and \textit{model-specific} methods - the first can be applied to any ML model, the second only to the ML model type it was developed for.

\subsubsection{Contrastive (counterfactual) explanations}

Contrastive explanations (CEs) are local, model-agnostic explanations.
They are computed \textit{after} the model was trained (post-hoc), and based only on input-output pairs.

Wachter et al.~\cite{DBLP:journals/corr/abs-1711-00399} introduced the idea of contrastive explanations in the field of XAI. Following them, a contrastive explanation can be described as an optimization problem of the following form:

\begin{equation} \label{equ:wachter}
    \arg \min_{x_{ce}} \max_{\lambda} \lambda(f_w(x_{ce})-y_{ce})^2 + d(x_f, x_{ce})
\end{equation}

$x_{ce}$ denotes the contrastive data point, $x_f$ the original data point, $f_w(x_{ce})$ the prediction of the contrastive example and $y_{ce}$ the desired prediction. Further, $\lambda$ denotes a tuning parameter, and $d(\cdot,\cdot)$ a distance measure, usually the Manhattan distance, weighted by the median absolute deviation. 
Equ. ~\ref{equ:wachter} minimizes the distance between the factual and the contrastive data point. 
The intuition underlying a CE is that to be realistic, it should be as close as possible to the original data point, while at the same time making sure that the prediction of the CE aligns with the desired prediction. 

Contrastive explanations are well received by the community.
However, they do not only have support from the technical side. While~\cite{DBLP:journals/corr/abs-1711-00399} is discussing the legal basis for contrastive explanations, with a focus on the European General Data Protection Regulation (GDPR),~\cite{DBLP:conf/ijcai/Byrne19} argues in favor of CEs from a psychological point of view. Further, both~\cite{DBLP:journals/ai/Miller19} and~\cite{DBLP:conf/fat/MittelstadtRW19} make clear that explanations in a contrastive form in general, i.e., explanations outlining why other events did not occur and differences to the actual outcome, are highly desirable for the (lay) end-user.
Finally, we point to a number of surveys:~\cite{DBLP:journals/corr/abs-2010-04050} focusing on the computational side of CEs and actionable recourse, a closely related field,~\cite{DBLP:conf/ijcai/KeaneKDS21}, critically analyzing the lack of evaluation methodology for CE methods, and two additional surveys~\cite{Guidotti2022,DBLP:journals/access/StepinACP21}.

\paragraph{Technically related approaches} 
The approach taken in this work is inspired by two strands of research, focusing either on solving CE queries by (a) using SAT, or causality based frameworks~\cite{DBLP:conf/aistats/KarimiBBV20,DBLP:conf/fat/KarimiSV21}
or (b) using ILP/MILP approaches
~\cite{DBLP:conf/kdd/CuiCHC15,DBLP:conf/ijcai/KanamoriTKA20,DBLP:conf/fat/Russell19,DBLP:conf/fat/UstunSL19}.
However, our approach is different in two main points: first, our focus is clearly on creating explanations for \textit{any} ML model. Approaches (a) are generally agnostic, but the model internals need to be known, in (b) only linear or additive models are considered. 
Second, to the best of our knowledge, this is one of the first approaches using CLP to generate CEs.

\subsubsection{Group contrastive (counterfactual) explanations}
The notion of group contrastive explanations is relatively new and refers to a small body of work that has not yet converged on a common definition.~\cite{DBLP:journals/corr/abs-2303-09297} proposes a definition of group contrastive explanations that is based on common \enquote{feature differences}, i.e., the idea that different but similar data points share the same countrastive example.
Their approach builds on a simple extension of the traditional approach for the generation of CEs such as~\cite{DBLP:conf/fat/MothilalST20}.
Further, the paper benchmarks its approach via a user study, showing its usefulness. 
Other works include
\cite{DBLP:journals/corr/abs-2205-08974}, focusing on explanations that refer to an ensemble of decisions, and
\cite{DBLP:conf/nips/RawalL20}, connecting group CEs to actionable recourse.

\subsubsection{Background knowledge}

Adding background knowledge to an explanation has the potential to significantly improve its quality~\cite{DBLP:journals/corr/abs-2105-10172,DBLP:conf/iclp/State21}. 
In this work, we refer to background (or prior) knowledge as to any information that is relevant in the decision context but that does not emerge through the decision pipeline.
Not only do simple facts count as such, but we would also consider a natural law as such knowledge, or as in our example in the introduction, a specific restriction a customer has related to her living reality (e.g., a minimum credit amount).
However, we have to restrict ourselves to knowledge that can be formalized and thus used by the explainer.

\reasonx{} incorporates knowledge in the form of linear constraints. 
Other examples of knowledge integration include the following.
A local explanation tool for medical data that incorporates an ontology (the ICD-9-CM) to generate a meaningful, local neighborhood for explanation generation~\cite{DBLP:conf/fat/PaniguttiPP20}. A discrimination discovery approach~\cite{DBLP:journals/tkdd/RuggieriPT10}, where knowledge in the form of association rules is used to detect cases of indirect discrimination.
Explanations are also historically connected to knowledge and logic reasoning - the first systems offered to explain AI models were expert systems~\cite{DBLP:journals/widm/ConfalonieriCWB21}.

\subsubsection{Interactivity}

Interactivity as property of an explanation aligns closely with our working definition: \enquote{an explanation, or explainability is about an interaction, or an exchange of information}, where it crucially matters to \textit{whom} the explanation is given, and for \textit{what} purpose~\cite{DBLP:conf/iclp/State21}.

While being acknowledged as an important property of an explanation that is successfully communicated to the end-user~\cite{DBLP:journals/ai/Miller19,DBLP:journals/corr/abs-1712-00547,DBLP:journals/cacm/WeldB19} - interactivity is only in very few cases incorporated into XAI methods.
Sokol and Flach~\cite{DBLP:journals/ki/SokolF20} outline the usefulness of interactivity prominently in their paper: \enquote{Truly interactive explanations allow the user to tweak, tune and personalise them (i.e., their content) via an interaction, hence the explainee is given an opportunity to guide them in a direction that helps to answer selected questions}
Also, they present a first solution (the glass-box tool~\cite{DBLP:conf/ijcai/SokolF18a}): an interactive explanation tool that provides explanations in natural language and that can be queried either by voice or via a chat. 
{Similar to our work, it relies on a decision tree to generate the explanations.}
The tool was tested in different environments, this helped the authors to also develop a mapping of desiderata of interpretable explanations.

We further point to~\cite{DBLP:journals/corr/abs-2202-01875}, presenting an interview study with practitioners that revealed that interactivity for explanations is strongly preferred, and discussing some ideas about how to achieve this.
As a last example, we point to the what-if tool, a commercial application provided by Google.\footnote{\href{https://pair-code.github.io/what-if-tool/}{https://pair-code.github.io/what-if-tool/}}

\subsection{(Constraint) Logic Programming}

Logic programming (LP) is a declarative approach to problem-solving, based on logic rules in the form of (Horn) clauses~\cite{Apt1997}. It supports reasoning under various settings, i.e., deductive, inductive, abductive and meta-reasoning~\cite{DBLP:journals/jair/CropperD22,10.5555/773294}. 
Starting with the Prolog programming language~\cite{DBLP:books/daglib/0076175}, programming in logic has been extended in several directions, as per expressivity and efficiency~\cite{DBLP:journals/tplp/KornerLBCDHMWDA22}. Constraint logic programming (CLP) augments logic programming with the ability to solve constrained problems in some domain~\cite{DBLP:journals/toplas/JaffarMSY92}. 
We rely on CLP($R$), which consists of linear constraints over the reals, as implemented in the SWI Prolog system~\cite{DBLP:journals/tplp/WielemakerSTL12}. CLP($R$) adds to logic programming rules the ability to test for linear constraint satisfiability, entailment, equivalence, and projection, as well as for solving MILP optimization problems.

\paragraph{(C)LP for XAI}

Logic programming is considered symbolic reasoning - contrary to what is commonly referred to as sub-symbolic reasoning, or ML. While the first set of approaches is inherently transparent, most ML models are not, but come with other advantages, such as the ability to work on large amounts of data. 
Combining both is a promising synergy - exactly what we do in this paper. 

A related approach to our work is a body of papers by Sokol et al. \cite{sokol2021intelligible,DBLP:conf/ijcai/SokolF18a,DBLP:journals/corr/abs-2005-01427}. While the first introduces explanations for decision trees (among others, through constrastive explanations), the second generalizes it to be used as a local surrogate model. In the third paper, the focus is explicitly set on interactivity.
Our work is closely linked to that. The main difference is in methodology, i.e., we rely on CLP in our computations, thus allowing for the integration of background knowledge, interactivity, and operation under under-specified information, making \reasonx{} unique against previous approaches.
Another related work is \cite{DBLP:journals/corr/abs-2011-07423}. It provides CEs - but treats them from the angle of (actual) causality. Methodologically, it relies on answer set programming (ASP), and uses also its straightforward ability to integrate knowledge into the explanations.

A survey that discusses how to combine both symbolic (logic) and sub-symbolic systems, with a specific focus on explainable AI is~\cite{DBLP:journals/ia/CalegariCO20}.
Further, \cite{DBLP:journals/corr/abs-2109-08290} adopts ASP to compute both local and global explanations for tree ensemble models.
While it is methodologically related to our approach (relying on logic programming and on tree structures), it does not discuss the notion of CEs.
There are also a few approaches using logic rules as explanations, but that methodologically do not rely on LP, 
e.g., \cite{DBLP:journals/expert/GuidottiMGPRT19} and \cite{DBLP:journals/ai/SetzuGMTPG21}.

\section{REASONX: Reason to explain}
\label{sec:reasonx}

We propose \reasonx{}, a novel tool to generate contrastive explanations that can account for background knowledge, that works under under-specified information and is highly interactive.
\reasonx{} can be used to 
obtain factual and contrastive decision rules about the classification of the data point (or profile) of interest, and provides the closest CE (via optimization).
In all settings, background knowledge can be added in the form of linear constraints.

\reasonx{} is strongly guided by a decision tree model, called the \textit{base model}. Such a decision tree can be: (a) the model to be explained and reasoned about; (b) a global surrogate model of a black box; (c) a local surrogate model of a black box decision in the neighborhood of an instance to explain. 
In cases (b) and (c), the surrogate model is assumed to have good fidelity in reproducing the black box decisions. This is reasonable for local models, i.e., in case (c), by learning the tree over a localized neighborhood as common in perturbation-based explanation methods such as LIME~\cite{DBLP:conf/kdd/Ribeiro0G16}. 
Regarding case (b), we point out that our approach works with any type of decision tree, including axis-parallel, oblique, and complete decision trees -- the last one offering very good performances.
For presentation purposes, we assume the case (a), in order to disentangle the properties of \reasonx{} in detail under no gap in fidelity between the surrogate model and the black box that should be explained.

The base model is translated into a set of Prolog facts, one for each path in the decision tree. In fact, a split in a decision tree is a linear constraint over features $\mathbf{x}$ of the form $c^T \mathbf{x} \geq b$ or $c^T \mathbf{x} < b$. A path from the root to a leaf is then a conjunction of linear constraints, represented by a Prolog fact such as:
\[ \mathit{path}(m,[\mathbf{x}], [c_1^T \mathbf{x} \geq b_1, \ldots, c_k^T \mathbf{x} \geq b_k], c, p).
\]
where $m$ is an id of the path, $[\mathbf{x}]$ the list of features, $c$ the class predicted at the leaf, $p$ the confidence of the prediction, and $[c_1^T \mathbf{x} \geq b_1, \ldots, c_k^T \mathbf{x} \geq b_k]$ the list of $k$ splits from the root to the leaf. Such linear constraints can be combined with constraints on data types (modeling for instance one-hot encoded features), on distance functions (modeling norms\footnote{\reasonx{} currently implements the L1 norm.} to be used in computing contrastive examples), and user-provided ones (the background knowledge). Further, a constraint $\varphi$ can be reasoned about in the following forms:
\begin{itemize}
    \item checking satisfiability, i.e., whether $\exists_{\mathbf{x}}\ \varphi$ holds, also considering some features from $\mathbf x$ over the domain of integers, e.g., the one-hot encoded features or ordinal features;
    \item projecting over some features $\mathbf{w} \subseteq \mathbf{x}$, i.e., computing $\exists_{\mathbf{x}\setminus\mathbf{w}}\ \varphi$, as a way to express $\varphi$  w.r.t. features in $\mathbf w$ only;
    \item checking entailment, i.e., whether they entail some other linear constraint, e.g., in order to test if solutions of $\varphi$ satisfy some property;
    \item solving a MILP minimization problem $\min f(\mathbf{x}) s.t. \varphi$ where $f(\cdot)$ is a linear function, e.g., in the distance minimization when computing contrastive instances.
\end{itemize}
The reasonings above are implemented through the CLP($R$) functionalities. However, to allow for wider accessibility of \reasonx{}, we build a Python layer on top of the CLP($R$) core, which translates the base model and the user queries into Prolog queries, and translates back the results into Python data structures. Python also adds some syntactic sugar to express more concisely a few categories of constraints, such as the immutability of features, or to generate constraints from the features of an instance.

\section{Synthetic Data Set Illustration} 
\label{sec:synthetic_dataset}

To illustrate the capabilities of \reasonx{}, we introduce a simple synthetic data set, comprised of two classes with each 1,000 data instances. These were sampled from different random normal distributions over two independent features. The distributions were generated to be almost separable by an axis-parallel decision tree.
We choose a data instance from class 0, called the \textit{factual instance} (``factual'' in the figures), to illustrate a few of the functionalities of \reasonx{}.

\begin{figure}[t]
    \centering
    \includegraphics[width=0.49\linewidth]{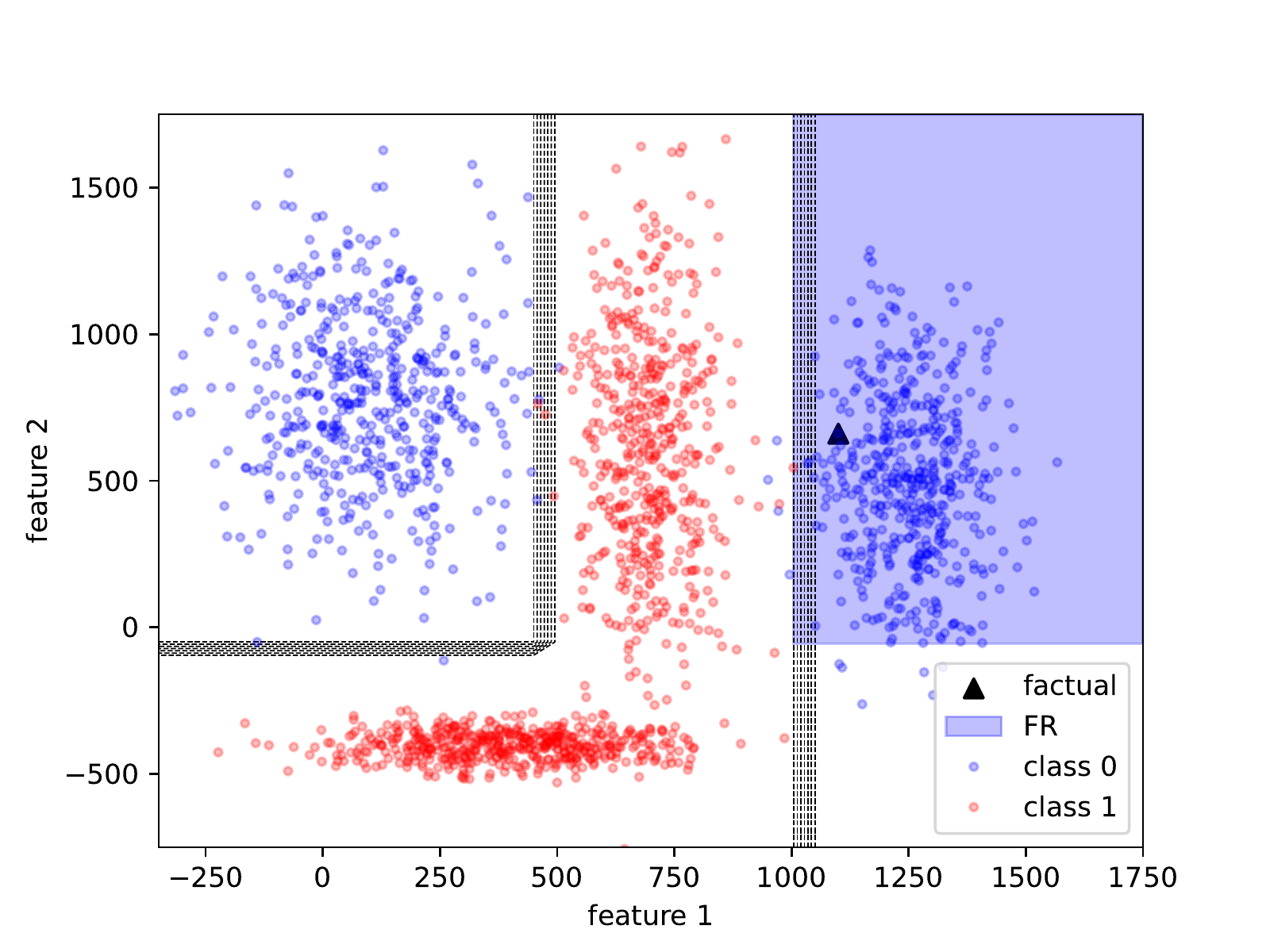}
    \includegraphics[width=0.49\linewidth]{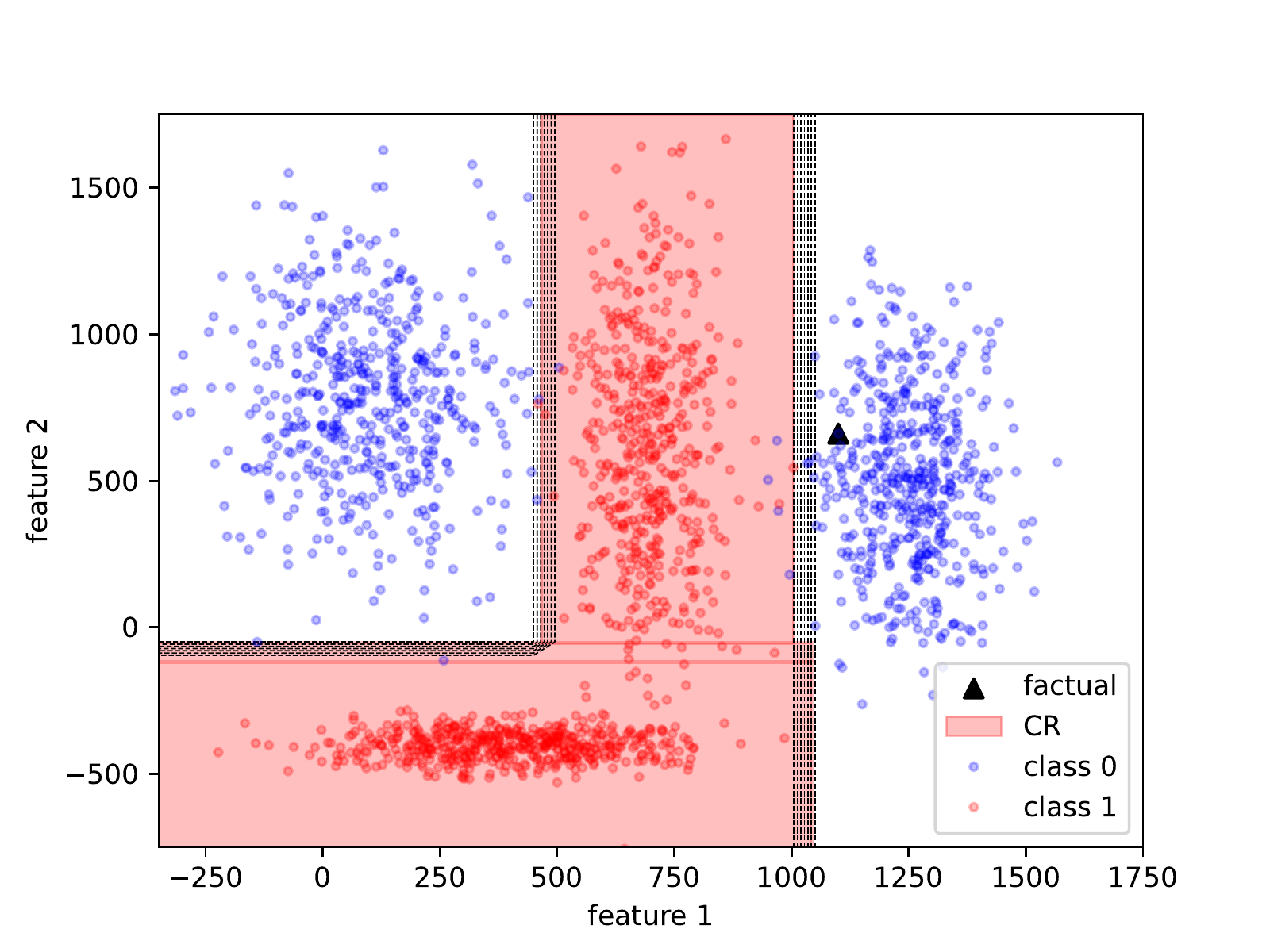}
    \caption{Left: factual region (FR) as provided by \reasonx{}. Right: contrastive regions (CR) as provided by \reasonx{}. Grey lines refer to the decision boundary of the tree.}
    \label{fig:factual_ce_simple}
\end{figure}

As a first operation, we use \reasonx{} to provide us with the factual rule (plotted as factual region in Fig. \ref{fig:factual_ce_simple}, left) that is satisfied by the data instance.
Next, we ask for the contrastive rules, i.e., admissible contrastive regions where potential contrastive examples are located (plotted as contrastive regions in Fig. \ref{fig:factual_ce_simple}, right). In this example, three of those regions exist.

\begin{figure}[t]
    \centering
    \includegraphics[width=0.49\linewidth]{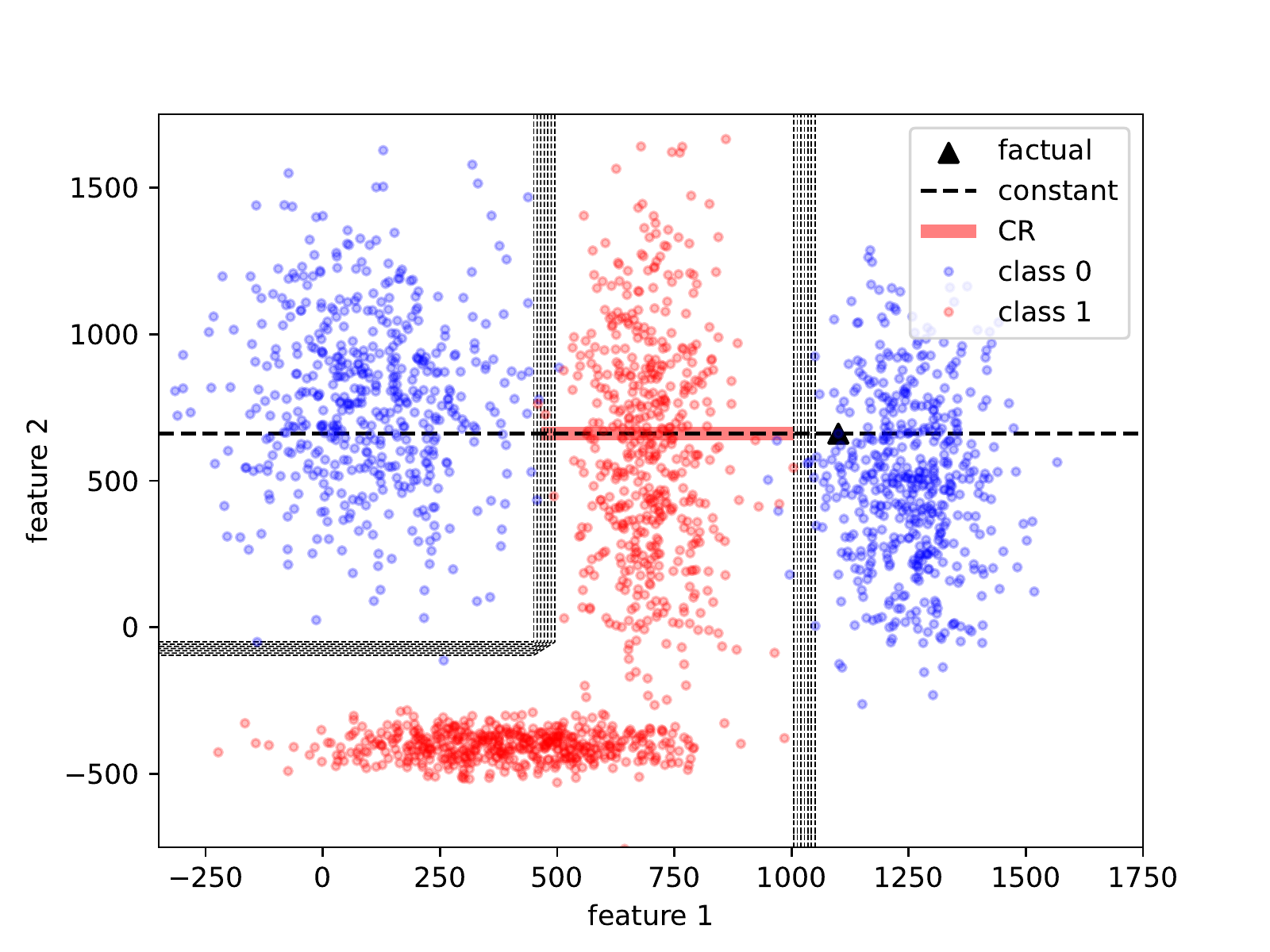}
    \includegraphics[width=0.49\linewidth]{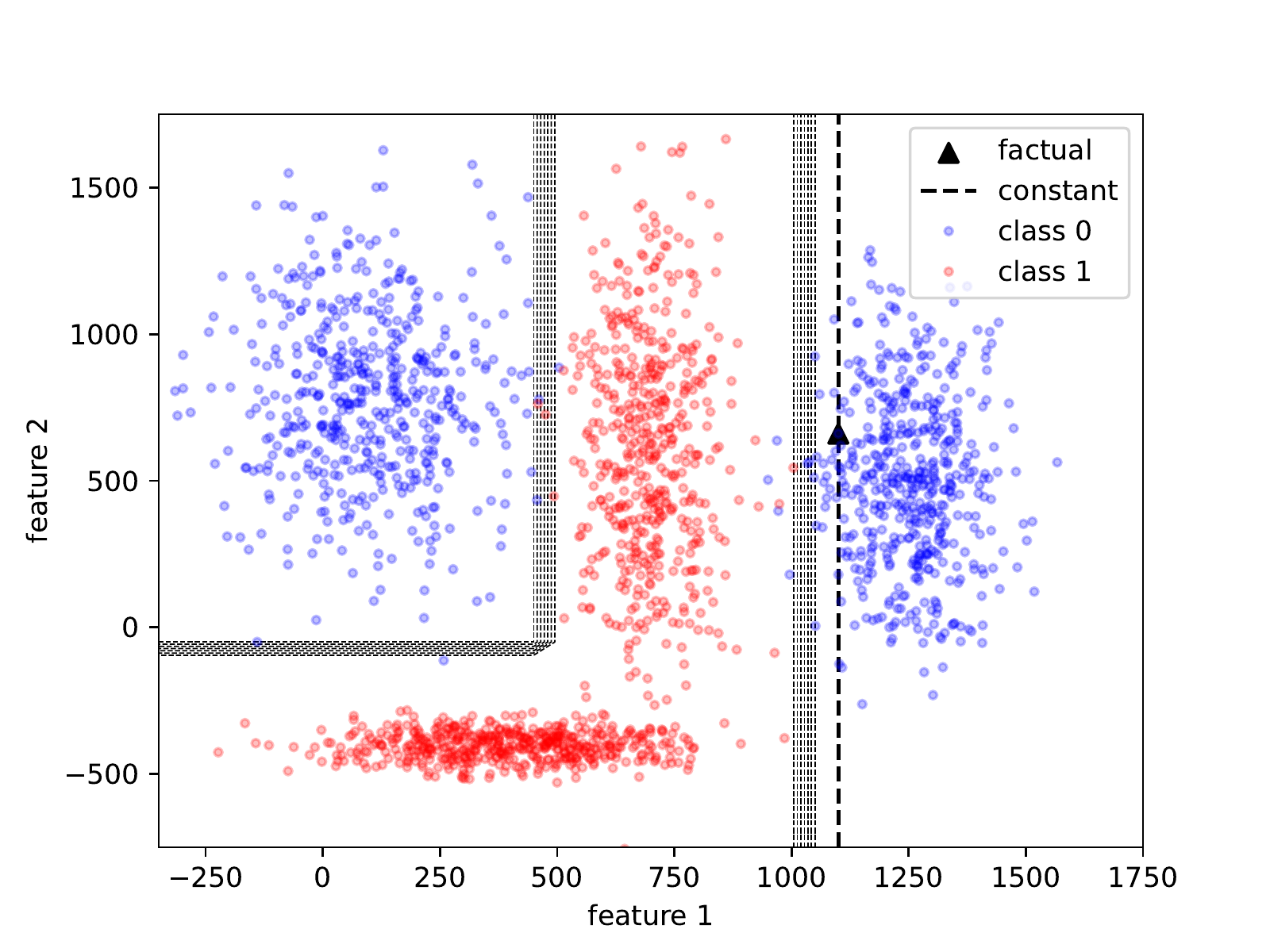}
    \caption{Left: contrastive region (CR) as provided by \reasonx{}, and given a constraint on feature 2. Right: given the constraint on feature 1, no solution provided by \reasonx{}. The dashed line refers to the enforced constraint. Grey lines refer to the decision boundary of the tree.}
    \label{fig:ce_constant}
\end{figure}

In Fig. \ref{fig:ce_constant} we demonstrate the capability of \reasonx{} to account for background knowledge. 
We use a constraint that ensures that feature 2 stays constant between the factual instance and the contrastive example (Fig. \ref{fig:ce_constant}, left), or a constraint that ensures that instead feature 1 stays constant (Fig. \ref{fig:ce_constant}, right). While the first leads to an admissible contrastive region in the form of a region (a line), the second constraint suppresses any solution (no solution exists).

\begin{figure}[t]
    \centering
    \includegraphics[width=0.49\linewidth]{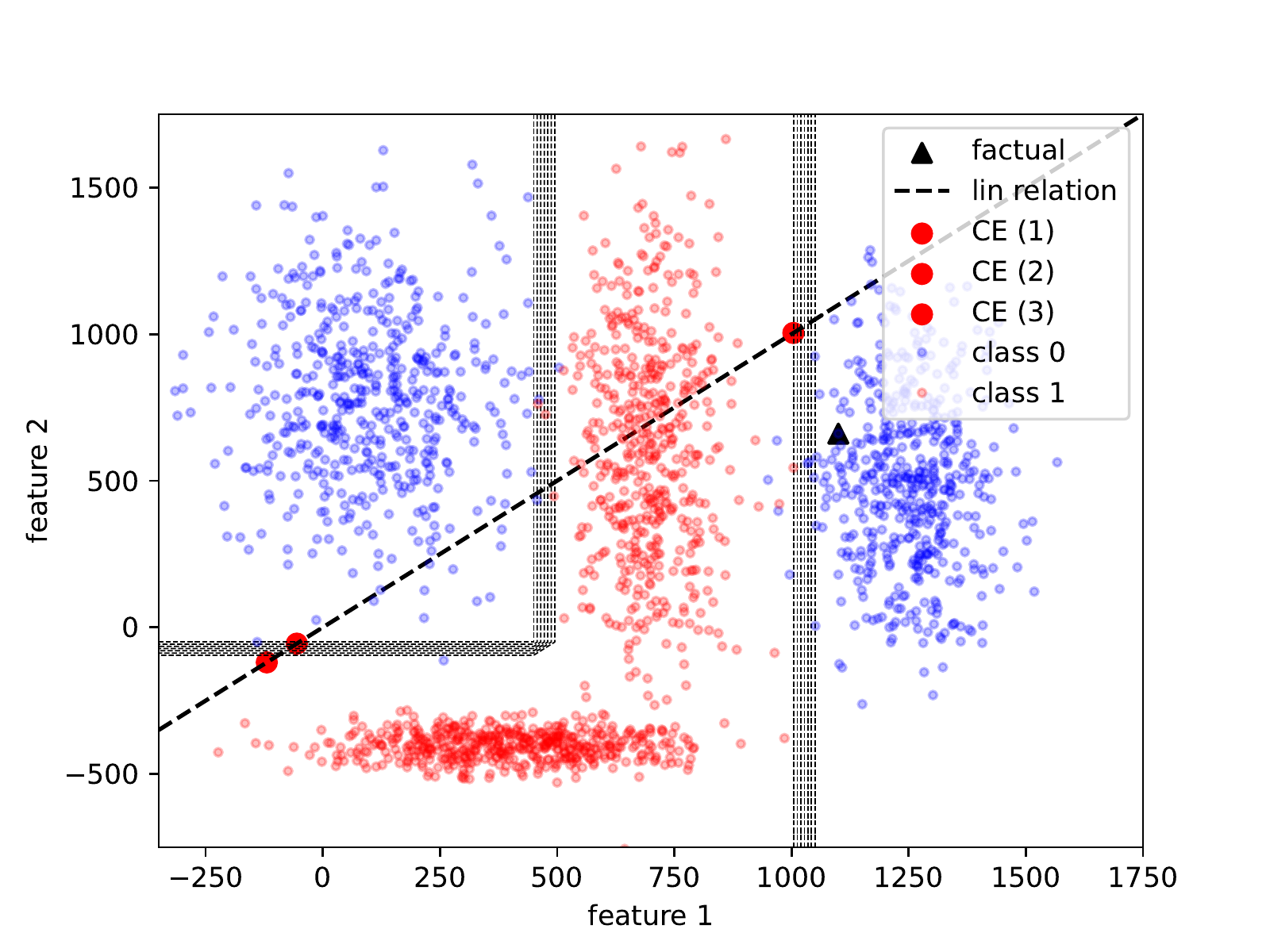}
    \includegraphics[width=0.45\linewidth]{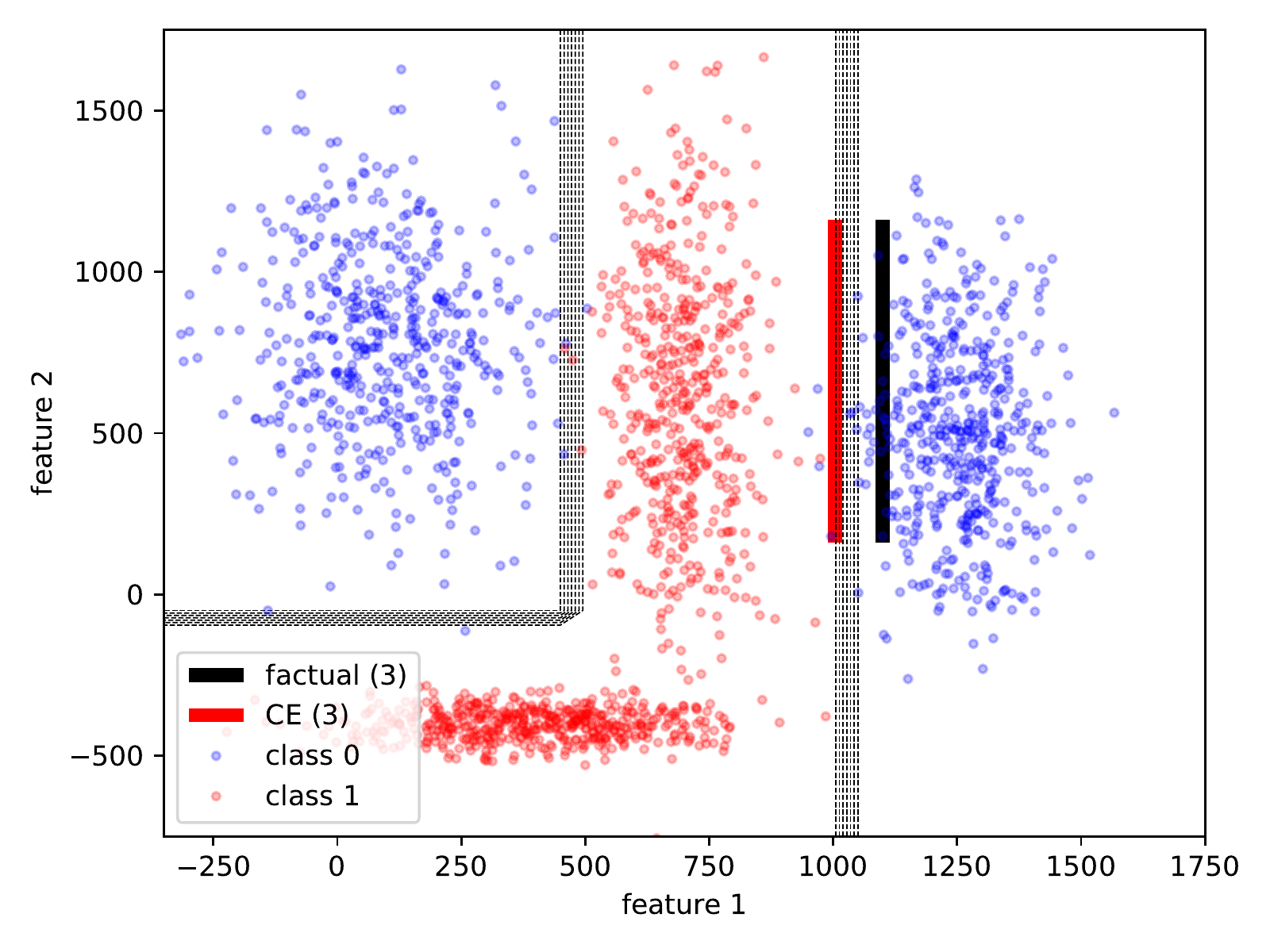}
    \caption{Left: closest CEs as provided by \reasonx{}, given a constraint on feature 1 and feature 2. The dashed line refers to the enforced constraint. Right: one closest CE as provided by \reasonx{}, feature 2 of factual is provided as region (solution 1). Grey lines refer to the decision boundary of the tree.}
    \label{fig:ce_closest_region}
\end{figure}

Now, we extend the example along two dimensions: asking for the \textit{closest} contrastive example under a specified constraint setting, and admitting under-specified information in the factual instance.
In Fig. \ref{fig:ce_closest_region}, left, we run \reasonx{} under a linear constraint between the two features (feature 1 and feature 2 have to be equal to each other). Instead of solving for admissible contrastive regions, we ask for the \textit{closest} CE. The solution is provided as red dots.
We observe the following: 
first, all solutions lie on the line that marks the linear constraint as introduced. 
Second, \reasonx{} provides not only one but three solutions with a different distance to the factual instance. 
This stems from the fact that in this example, three admissible contrastive regions (see also Fig. \ref{fig:factual_ce_simple}, right) are given. \reasonx{} solves the optimization for each of those independently, and provides us therefore not with the global optimum (one solution), but three local optima. 
This can be an advantage over conventional tools that provide contrastive explanations, as it leaves more flexibility to the end-user. A further refinement of the results, e.g. filtering for the global optimum, can be implemented.

\begin{figure}[h]
    \centering
    \includegraphics[width=0.49\linewidth]{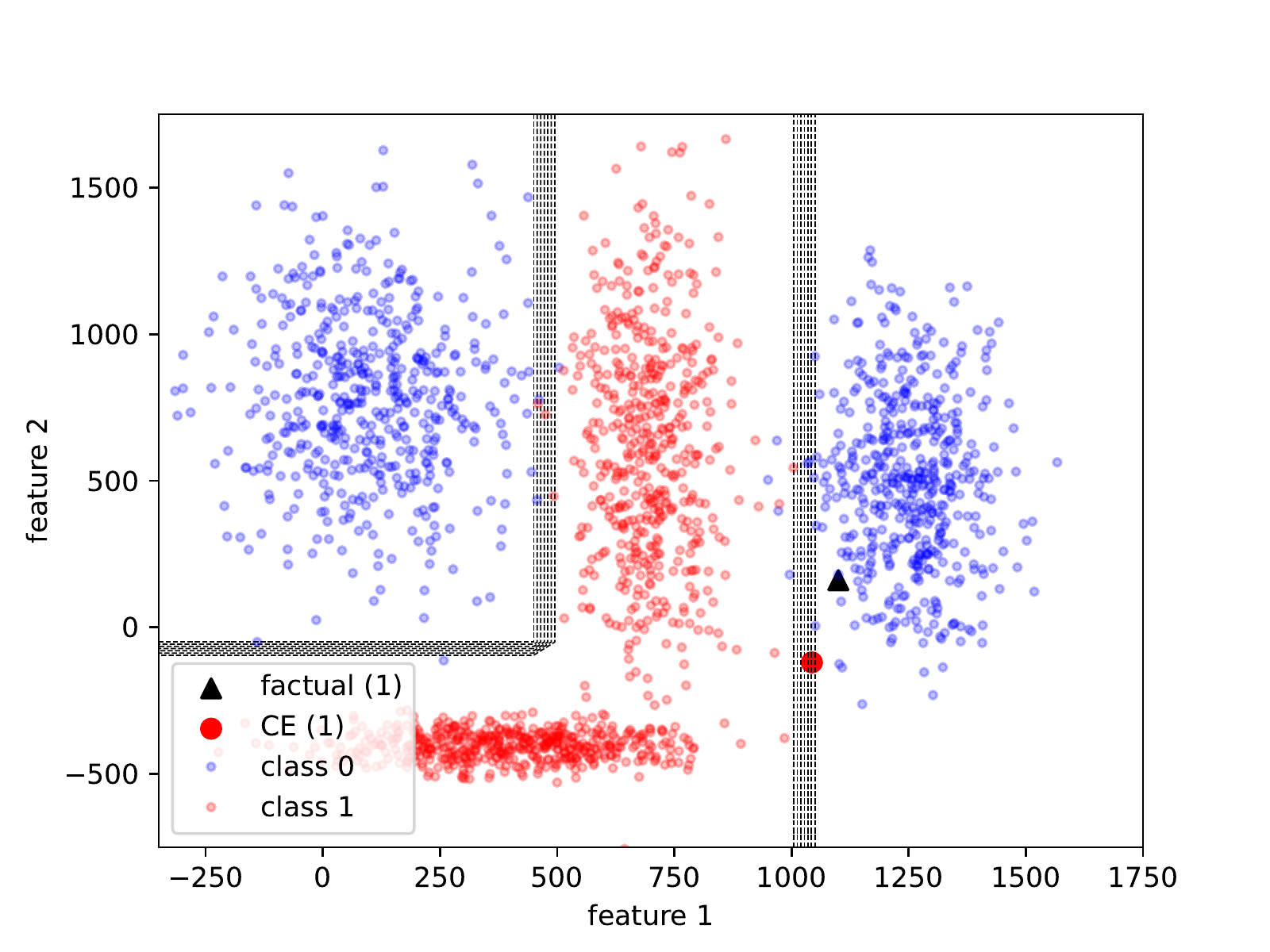}
    \includegraphics[width=0.49\linewidth]{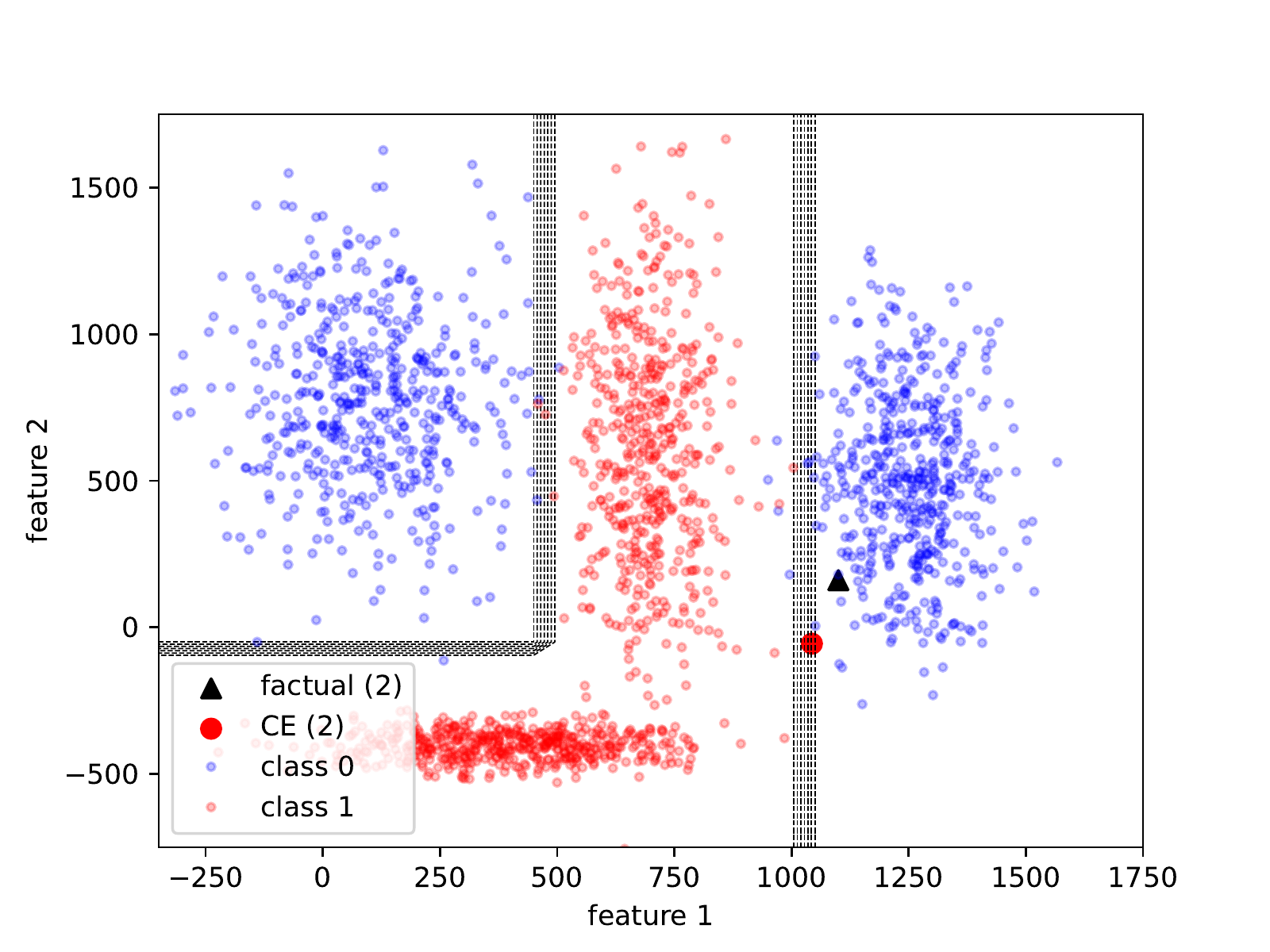}
    \caption{Closest CE as provided by \reasonx{}, feature 2 of factual is provided as region (solution 2 and 3).
    Grey lines refer to the decision boundary of the tree.}
    \label{fig:ce_condition5}
\end{figure}

In Fig. \ref{fig:ce_closest_region}, right, we re-initialize the factual instance and allow feature 2 to be under-specified, i.e., setting it to a region instead of a fixed value. We omit all constraints, and solve again for the \textit{closest} CE. We display one solution of this query in Fig. \ref{fig:ce_closest_region}, right. For the same reasons as above, there exist three solutions (other two solutions in Fig. \ref{fig:ce_condition5}).
We observe that the provided solution is also a region in the data space.

\section{Credit Application Example}
\label{sec:credit_use_case}

This example focuses on contrastive explanations for an ADM system that assesses credit applications, for example, provided by a bank.
The main \textit{purpose} of the explanation is for a customer (our imagined end-user) to make sense of the decision, and to receive insights about possible opportunities of action.
Secondary purposes are legal compliance of the ADM, as well as increasing the trust of its customers in internal decisions, both in the interest of the bank.
Next to demonstrating the capabilities of \reasonx{}, this example serves as a (qualitative) \textit{validation} of the tool. 
This is justified by the strong qualitative nature of \reasonx{}, and ties closely to the fact that an explanation tool always has to be discussed in context~\cite{DBLP:journals/corr/abs-1901-04592}.

We run the example on a processed version of the Adult Income data set\footnote{\href{https://archive.ics.uci.edu/ml/datasets/Adult}{https://archive.ics.uci.edu/ml/datasets/Adult}}, which has a binary response variable.
It contains twelve features, and we restrict ourselves to a subset: three categorical features (\texttt{workclass}, \texttt{race}, \texttt{sex}), one ordinal (\texttt{education}) and four continuous (\texttt{age}, \texttt{capitalgain}, \texttt{capitalloss}, \texttt{hoursperweek}).
Categorical features are one-hot encoded. 
The queries as displayed below work iteratively, i.e., background knowledge that is added to the program is always kept (but can be retracted).
The \texttt{verbose} parameter determines how the output of the query is displayed.
The \texttt{project} parameter is used to map results of the set of variables onto a subset.

After initializing \reasonx{} with the base decision tree trained on a subset of available data, we turn towards a first question of the user (corresponding to the factual instance): \textit{Why was my application rejected?} We answer it
by showing the rule behind the decision.
This requires naming an instance (\texttt{'F'} in the code below, for `factual') and passing the feature values (\texttt{InstFeatures}) and the decision of the base model (\texttt{InstDecision}). 
\begin{small}
\begin{mdframed}
\begin{verbatim}
USER:    r.instance('F', features=InstFeatures, label=InstDecision)
         r.solveopt(verbose=2)
REASONX: Rule satisfied by F: 
         IF F.capitalgain<=5119.0,F.education<=12.5,F.age<=30.5 
         THEN <=50K [0.9652]
\end{verbatim}
\end{mdframed}
\end{small}
The rule as returned by \reasonx{} explains the classification of the factual instance. It refers to a specific region in the data input space as characterized by the base decision tree.
A second answer can be given by comparing the factual instance against a contrastive rule, using the differences as an explanation. This requires naming a second instance (\texttt{'CE'} in the code below, for `contrastive example'), and possibly a minimum confidence of the rule leading to the contrastive decision.
We obtain two rules by running the following query:
\begin{small}
\begin{mdframed}
\begin{verbatim}
USER:    r.instance('CE', label=1-InstDecision, minconf=0.8)
         r.solveopt(verbose=2)
REASONX: Rule satisfied by CE: 
         IF CE.capitalgain>5119.0,CE.capitalgain<=5316.5 
         THEN >50K [1.0],
         Rule satisfied by CE: 
         IF CE.capitalgain>7055.5,CE.age>20.0 
         THEN >50K [0.9882]
\end{verbatim}
\end{mdframed}
\end{small}
By comparing this output with the answer to the previous question, the user can understand the factual decision of the ADM better.
This is especially relevant to answer a second question: \textit{What are my options to change the outcome of the ADM, and receive the credit?} 
For example, comparing the first contrastive with the factual rule, an increase of the feature \texttt{capitalgain} will lead to a change in the predicted outcome of the ADM. 
Similarly, an increase in the \texttt{capitalgain}, and a change in the \texttt{age} (from $19$ to $20$ or higher) will alter it.

In the next step, we add some background knowledge to the explanations. 
We apply an immutability constraint on the feature \texttt{age}:
\begin{small}
\begin{mdframed}
\begin{verbatim}
USER:    r.constraint('CE.age = F.age')
         r.solveopt(verbose=2)
REASONX: Rule satisfied by CE: 
         IF CE.capitalgain>5119.0,CE.capitalgain<=5316.5 
         THEN >50K [1.0]
\end{verbatim}
\end{mdframed}
\end{small}
As expected, the solution has changed: by adding the above-stated constraint, the admissible region for CEs becomes smaller, and only one contrastive rule remains.
Last, we ask for the CE that is \textit{closest} to the factual instance: 
\begin{small}
\begin{mdframed}
\begin{verbatim}
USER:    r.solveopt(minimize='l1norm(F, CE)',  project=['CE'], 
         verbose = 2)
REASONX: Answer constraint:
         CE.race=Black, CE.sex=Male,
         CE.workclass=Private,
         CE.education=10.0,
         CE.age=19.0,
         CE.capitalgain=5119.0,
         CE.capitalloss=0.0,
         CE.hoursperweek=40.0
\end{verbatim}
\end{mdframed}
\end{small}
\reasonx{} returned the \textit{closest} CE. This corresponds to the most common notion of a CE in the literature. However, \reasonx{} also took care of the specified background knowledge - the CE is returned under the above specified constraints on the feature \texttt{age}.
This is a similar case as described in Sec. \ref{sec:synthetic_dataset}, Fig. \ref{fig:ce_closest_region}, left. 
We extend the example to account for under-specified information. E.g., it can be interesting to ask for the \textit{closest} CE in the case the feature \texttt{age} is not fixed, but restricted to $19$ years or lower. 

\begin{small}
\begin{mdframed}
\begin{verbatim}
USER:    r.retract('F.age=19.0')
         r.constraint('F.age<=19.0')
         r.solveopt(minimize='l1norm(F, CE)', project=['CE', 'F.age'],
         verbose = 2)
REASONX: Answer constraint:
         CE.race=Black, CE.sex=Male,
         CE.workclass=Private,
         CE.education=10.0,
         CE.age=F.age,
         CE.capitalgain=5119.0,
         CE.capitalloss=0.0,
         CE.hoursperweek=40.0
\end{verbatim}
\end{mdframed}
\end{small}

The returned solution of \reasonx{} is similar to the previous one, but we observe that also in the CE, the feature \texttt{age} is not fixed to a single value but is provided as an equality region. This is a nice demonstration of how both in the input and the output, \reasonx{} does work under under-specified information. 
For a similar example based on the synthetic data set, see Sec. \ref{sec:synthetic_dataset}, Fig. \ref{fig:ce_closest_region}, right.

Since queries work iteratively, the flow of the above corresponds exactly to how an \textit{interaction} with \reasonx{} could look like, forming what we call an explanation (dialogue). 
Repeated specification of background knowledge and querying are a central part of this dialogue between the end-user and \reasonx{}, allowing to build individual, personalized explanations.

\section{Limitations}
\label{sec:limitations}

\paragraph{Evaluation}

So far, we demonstrated the capabilities of \reasonx{} via a synthetic illustration and an example in the credit domain, and - by surveying relevant literature - we showed how \reasonx{} is novel against other XAI tools.
The most prominent difference between \reasonx{} and other XAI methods is that it relies on symbolic reasoning, based on CLP capabilities, to generate (contrastive) explanations. This is exactly what makes it possible for \reasonx{} to combine a set of useful properties: background knowledge integration, interactivity, and operation under under-specified knowledge.
However, a thorough  evaluation of \reasonx{} is currently \textit{missing} from this paper.
While there is yet no fixed consensus on how to evaluate an explanation, in the case of \reasonx{}, a qualitative and quantitative evaluation is especially tricky: no method exists so far for explanation tools utilizing CLP.
Developing a proposal of a set of suitable methods is therefore a priority for future work.
We also point out the possibility of user studies for evaluation~\cite{DBLP:journals/corr/abs-2210-11584}, or validation via real-world data~\cite{DBLP:journals/corr/abs-1901-04592}.

\paragraph{Technical}We demonstrated the capabilities of \reasonx{} on (nominal, ordinal, and continuous) tabular data, and for a binary decision problem. While an extension to multi-class problems is straightforward, the extension to unstructured data, such as text and images, needs to be strictly formalized.
In the domain of images, a solution can be the integration of concepts, as demonstrated by~\cite{DBLP:conf/aiia/DonadelloD20}.

\paragraph{Social}

We discussed the delivery of explanations to enable the end-user to understand the decision better. Another reason to compute explanations is to understand algorithmic harms such as bias.
An important question is: \textit{Is the protected attribute (e.g., gender, race, age \cite{EU}) determining the decision outcome?} The answer can be given by e.g., comparing a factual instance and the CE, with a focus on whether only protected attributes alter the decision outcome.

Another important point to consider is the fairness of explanations. For two popular model-agnostic, local methods it has been shown that explanations can be unfair: different fidelity values can apply for different subgroups \cite{DBLP:conf/fat/BalagopalanZHHR22}.

Last, we would like to discuss two basic assumptions of \reasonx{} that \textit{must} be taken care of in social application contexts.
We do not explicitly talk about recommendations based on contrastive explanations but must consider this as a secondary application, as discussed under the keyword of \textit{algorithmic recourse}~\cite{DBLP:journals/corr/abs-2010-04050}.
First, we assume that our model is \textit{stable over time}, making it possible not only to obtain several contrastive explanations but proposing that the CE suggests a change that indeed alters the outcome of the ML model.
While this might hold in a toy setting, it likely does not in practice and can create some wrong promises~\cite{DBLP:conf/fat/BarocasSR20,DBLP:conf/iclp/State21}.
Second, by querying the \textit{closest} CE, there is often the implicit assumption that the closer such a CE, the easier the proposed change. This does not necessarily hold in practice, as pointed out already~\cite{DBLP:journals/corr/abs-2010-04050}.
While an extension of \reasonx{} to account for a variety of norms in the optimization is possible and part of future work, we acknowledge the importance of the choice of a suitable distance function, depending on the context of the application.

\section{Conclusion}
\label{sec:conclusion}

We presented \reasonx{}, an explanation tool that interactively provides contrastive explanations that can account for background knowledge in the form of constraints, and that works under under-specified information.
Background knowledge has the potential to significantly improve the quality of an explanation, while interactivity is an important aspect of explanations to be communicated to the end-user. Further, our tool works on under-specified information (profiles) which leads to increased flexibility.
\reasonx{} provides explanations for decision trees, which can be the ML models under analysis, or global/local surrogate models of any ML model.
We demonstrated \reasonx{} both on a synthetic data set, and an example in the credit domain.

We aim at extending \reasonx{} along several directions: 
$i)$ the implementation of more constraints,
$ii)$ an extension to account for more expressive trees such as oblique/optimal decision trees, 
$iii)$ the application to local explanations.
Implementing other constraints (e.g., diversity of a set of CEs, or sparsity of changes from the factual instance to the CE) will lead to a more flexible tool, and possibly better explanations, depending on the context that will necessitate these constraints.
Relying on oblique/optimal decision trees instead of standard decision trees that use axis-parallel splits will allow to better fit the model onto the data, for higher accuracies, and therefore improved quality of the explanations, while not departing from reasoning over linear constraints. 
Last, \reasonx{} can be easily adapted as a local explanation tool. The necessary changes include learning the tree not over the input data but a local neighborhood, which can, for example, be generated using random perturbations as in LIME~\cite{DBLP:conf/kdd/Ribeiro0G16}.

We would also like to refer to the option to extend \reasonx{} in a direction that allows a tighter integration between symbolic and sub-symbolic approaches. One possibility is an extension that allows \reasonx{} to extract the rules which are fundamental to \reasonx{} not via a tree, but directly from an underlying sub-symbolic structure.

\paragraph{Acknowledgments.}
Work supported by the European Union’s Horizon 2020 research and innovation programme under Marie Sklodowska-Curie Actions for the project NoBIAS 
(g.a. No. 860630), and under the Excellent Science European Research Council (ERC) programme  for the XAI project (g.a. No. 834756).
This work reflects only the authors’ views and the European Research Executive Agency (REA) is not responsible for any use that may be made of the information it contains.


%
%

%
%
%

\newpage
\bibliographystyle{splncs04}
\bibliography{library}

\end{document}